\documentclass{article}

\usepackage{microtype}
\usepackage{graphicx}
\usepackage{subfigure}
\usepackage{booktabs} 
\usepackage{url} 

\usepackage[accepted]{icml2019}

\icmltitlerunning{Uncertainty estimates for soil moisture deep learnining}

\begin{document}

\twocolumn[
\icmltitle{Evaluating aleatoric and epistemic uncertainties of time series deep learning models for soil moisture predictions}



\icmlsetsymbol{equal}{*}

\begin{icmlauthorlist}
\icmlauthor{Kuai Fang}{ce}
\icmlauthor{Chaopeng Shen}{ce}
\icmlauthor{Daniel Kifer}{cs}
\end{icmlauthorlist}

\icmlaffiliation{ce}{Department of Civil and Environmental Engineering, Pennsylvania State University, University Park, Pennsylvania, USA.}
\icmlaffiliation{cs}{Department of Computer Science and Engineering, Pennsylvania State University, University Park, Pennsylvania, USA.}

\icmlcorrespondingauthor{Chaopeng Shen}{cshen@engr.psu.edu}

\icmlkeywords{Machine Learning, ICML}

\vskip 0.3in
]



\printAffiliationsAndNotice{\icmlEqualContribution} 

\begin{abstract}
Soil moisture is an important variable that determines floods, vegetation health, agriculture productivity, and land surface feedbacks to the atmosphere, etc. Accurately modeling soil moisture has important implications in both weather and climate models. The recently available satellite-based observations give us a unique opportunity to build data-driven models to predict soil moisture instead of using land surface models, but previously there was no uncertainty estimate. We tested Monte Carlo dropout (MCD) with an aleatoric term for our long short-term memory models for this problem, and asked if the uncertainty terms behave as they were argued to. We show that the method successfully captures the predictive error after tuning a hyperparameter on a representative training dataset. We show the MCD uncertainty  estimate, as previously argued, does detect dissimilarity. 
\end{abstract}

\section{Background}
\label{submission}

Soil moisture ($\theta$), quantified by the volumetric fraction of soil occupied by water, critically controls various environmental and ecosystem processes such as photosynthesis, evapotranspiration, runoff, soil respiration, flooding, land-atmosphere interactions \citep{Koster2004}, etc. For climate and weather modeling, soil moisture is typically provided by land surface models (LSM). However, LSMs often introduce bias. For example, LSM-simulated moisture tends to be high-biased in the arid western CONUS and low-biased in wetter eastern US \citep{Yuan2017}, or low-biased in wet seasons and high-biased in dry seasons \citep{Xia2015}. Prevalent bias with LSM-simulated moisture can introduce large errors to downstream applications including weather and climate modeling \citep{Massey2016}. 

To reduce such bias, recent work introduced time series deep learning (DL) models to learn soil moisture dynamics directly from satellite-based observations. Satellites like Soil Moisture Active Passive (SMAP) mission \citep{Entekhabi2010} now provides near real time monitoring of surface soil moisture and such data provide an opportunity for data-driven models. It now becomes possible to replace certain parts of the LSM functionality using machine learning predictions. Our previous work employed long short-term memory (LSTM) to predict SMAP soil moisture, given meteorological forcings (precipitation, temperature, radiation, etc) \citep{Fang2017}. They showed that LSTM can extend SMAP to spatiotemporally seamless coverage of continental US (CONUS) with high fidelity to SMAP. In addition, inter-annual trends of root-zone soil moisture were surprisingly well captured by LSTM even when the model was trained using only three years of data \citep{Fang2018}.

Despite such progress, few of these studies, or any studies utilizing LSTM in the field hydrology to our knowledge, addressed model uncertainties. Uncertainty is critical for understanding the limitations and data needs of the model \citep{Pappenberger2006}. In the context of data-driven modeling, uncertainty is often regarded as a combination of two elements:  aleatoric and epistemic uncertainties. Aleatoric uncertainty is due to inherent stochasticity in the system. Epistemic uncertainty stems from our lack of knowledge (insufficient or biased training data) that could, in principle, be known. 

In this paper we test a recently-proposed uncertainty quantification (UQ) framework for DL models: Monte Carlo dropout (MCD) with an aleatoric uncertainty term \citep{Kendall2017}. It was argued that doing Monte Carlo dropout with a neural network is equivalent to doing variation Bayesian inference in a deep Gaussian Process \citep{Gal2016}. However, few studies showed definite proof that the MCD detects similarity in the input space. Due to the approximate derivations in their work, we must ask (i) whether this scheme is truly successful at predicting errors; and (ii) do the terms behave as proposed, i.e., does the epistemic term measures similarity to the training data as a Gaussian Process does.

\section{Method}
\label{submission}
MCD with an uncertainty term simultaneously estimates aleatoric and epistemic uncertainty. The MCD part measures disagreement among members in an ensemble of models generated by applying dropout \citep{Srivastava2014}. \citet{Gal2016} proposed that dropout training of deep networks was an approximation of training Gaussian Processes. Hence, we used dropout during prediction to create random ensemble, and used the variability of these predictions to quantify the epistemic uncertainty ($\sigma_{mc}$). The second is an input-dependent heteroscedastic model for observational noise \citep{Kendall2017} where a second output unit ($\sigma_{x}$) can be added to a deep network. With a specially-designed loss function, $\sigma_{x}$ can represent an estimate of the variance of the network's prediction. Finally we combined those two parts of uncertainty as $\sigma_{comb}^2 = \sigma_{mc}^2 + \sigma_{x}^2$. 
The MCD uncertainty estimates require calibration by adjusting a hyper-parameter. Our model training period was the first year. To avoid over-tuning, we searched for a uniform drop out rate value so that the uncertainty in the second year matches the magnitude of the actual error during that period. We reported metrics from the test period (the third year).

\section{Results and Discussion}
\begin{figure}[h]
\centering
\includegraphics[width=1\linewidth]{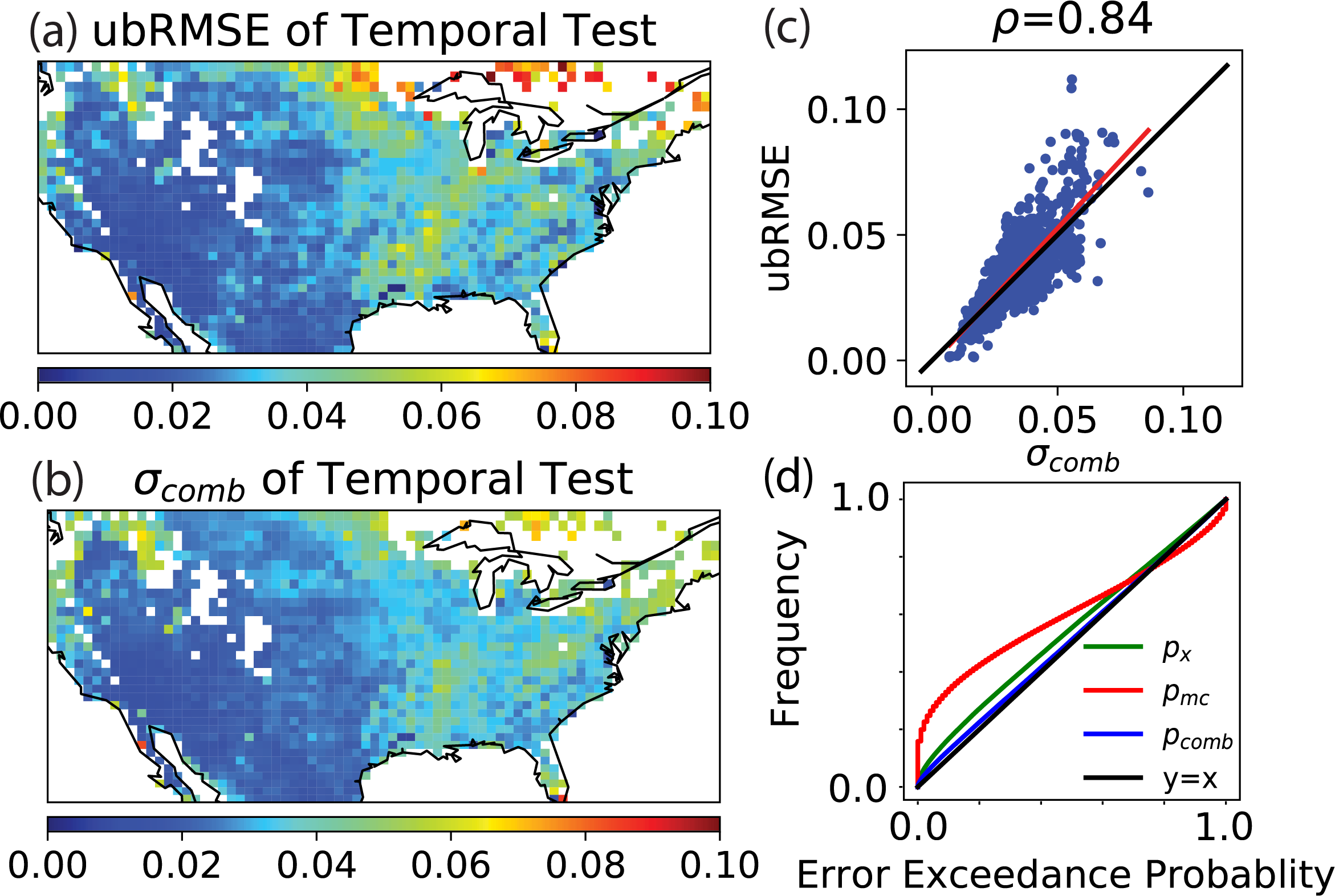}
\caption{Model error and uncertainty estimates of temporal generalization test (train in one year, tune the dropout rate in the second, and test in the third year) over CONUS. (a) unbiased root-mean squared error (ubRMSE), (b) $\sigma_{comb}$, (c) pixel-based comparison of ubRMSE vs. $\sigma_{comb}$, and (d) the calibration curve of the uncertainty estimates. Perfect uncertainty estimate is a straight one-to-one line.}
\label{fig_map_CONUS}
\end{figure}

To evaluate the overall quality of the uncertainty estimation, we trained a LSTM model to learn SMAP soil moisture dynamics for the entire CONUS using one year of data, and ran temporal test on another year. After tuning with the validation period, we chose a dropout rate of 0.6. As Figure \ref{fig_map_CONUS}a-b shows, the spatial pattern of $\sigma_{comb}$ agreed more or less with the predictive error (quantified by the unbiased root-mean-squared error, ubRMSE), and were larger in the East than in the West. SMAP signal is adversely impacted by large VWC \citep{ONeill2016} and freezing conditions further reduces the amount of training data for surface soil moisture \citep{Fang2018}. As a result, the Northeastern and Northwestern forests (along the Rocky mountains) had the highest ubRMSE. The lowest errors were found on the Great Plains and Southeast due to aridity and reduced forest cover. The predicted $\sigma_{comb}$ automatically captured these spatial patterns. This good performance is also evident from the high correlation between the $\sigma_{comb}$ (Figure \ref{fig_map_CONUS}c) and the nearly straight line in the calibration curve (the green line in Figure \ref{fig_map_CONUS}d). These results suggest that, for temporal prolongation, it is possible to anticipate model predictive errors using $\sigma_{comb}$.


Do the two uncertainty terms behave as asserted, i.e., does the aleatoric term respond to stochasticity in the data and does the epistemic term respond to dissimilar cases? To answer this question, we examined the behaviors of the MCD uncertainty estimates when models are trained on a small basin and test on other regions. 


\begin{figure}
\centering
\includegraphics[width=0.95\linewidth]{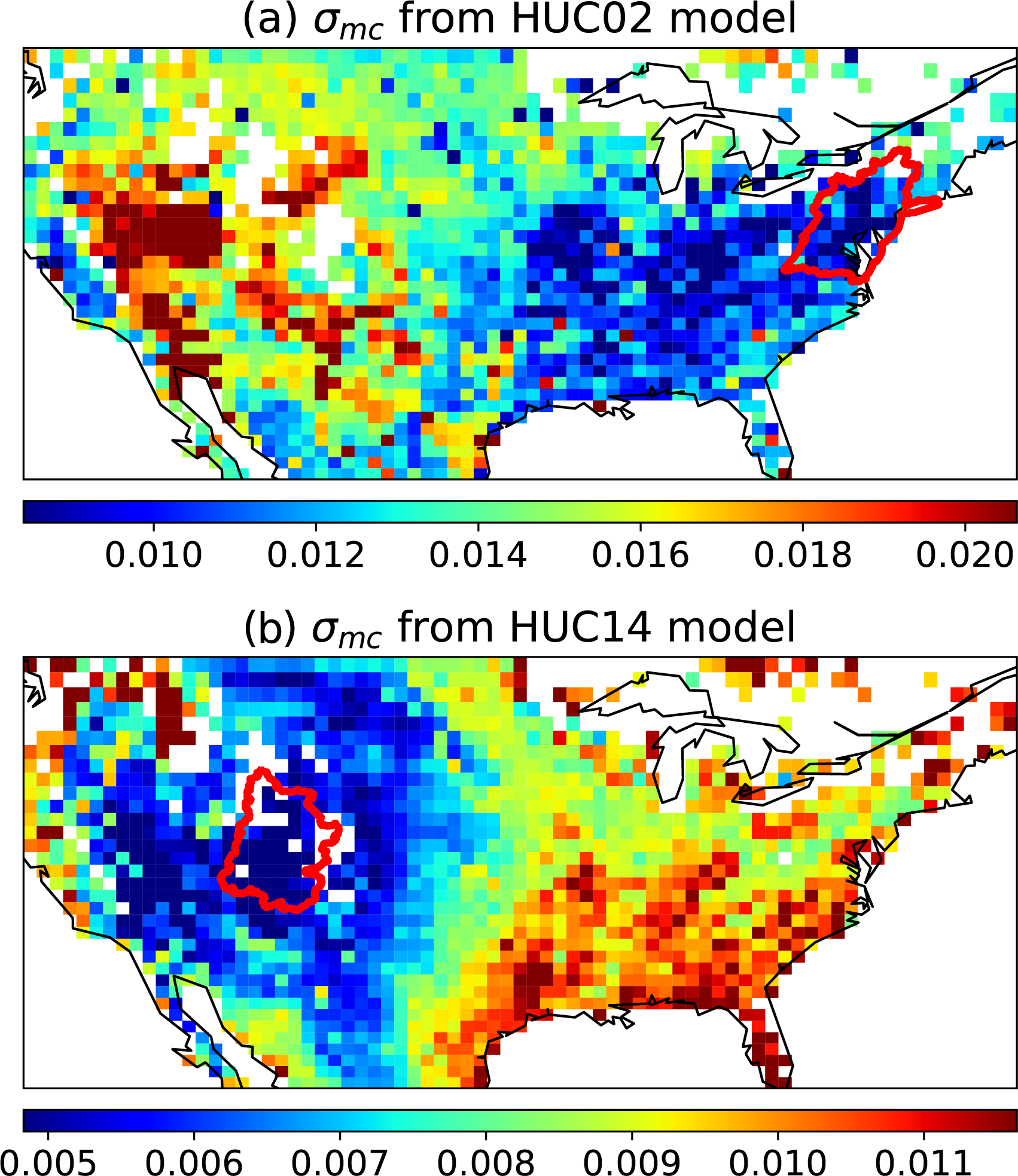}
\caption{Maps of $\sigma_{mc}$ when the LSTM model is trained in one of the HUC2 basins. The training region is highlighted by the red polygon.}
\label{fig_sigmaMC_GP}
\end{figure}

We trained models on each of the 18 level-2 hydrologic cataloging unit (HUC2) basins dividing CONUS and we show the $\sigma_{mc}$ when the model is tested in other regions. $\sigma_{mc}$ is the smallest inside the training region, somewhat larger on the neighboring region, and much larger further away (Figure \ref{fig_sigmaMC_GP}). This result provides a clear and novel evidence that MCD does detect dissimilarity in the input space, which manifest as geographic proximity in this case. Note our inputs do not have any attribute that directly represents location, and therefore the sense of proximity or similarity was discovered by the network itself via lands surface characteristics such as soil texture and climatology, which are auto-correlated. 

In summary, we evaluated the suitability of the MCD method with an aleatoric term for hydrologic datasets, using SMAP-based soil moisture product as a test case. Our evaluation shows that the proposed scheme can be effective at predicting the model error after tuning the dropout rate. Our results provide a unique and strong evidence that variational sampling via Monte Carlo dropout acts as a dissimilarity detector. The aleatoric term was also found to be effective. 





\bibliography{example_paper}
\bibliographystyle{icml2019}



\end{document}